# "*Pale as death*" or "*pâle comme la mort*": Frozen similes used as literary clichés

**Suzanne Mpouli**
LIP6, Université Pierre et Marie Curie
Labex OBVIL, Paris IV
mpouli@acasa.lip6.fr

**Jean-Gabriel Ganascia**
LIP6, Université Pierre et Marie Curie
Labex OBVIL, Paris IV
jean-gabriel.ganascia@lip6.fr

## Abstract

The present study is focused on the automatic identification and description of frozen similes in British and French novels written between the 19$^{th}$ century and the beginning of the 20$^{th}$ century. Two main patterns of frozen similes were considered: adjectival ground + simile marker + nominal vehicle (e.g. *happy as a lark)* and eventuality + simile marker + nominal vehicle (e.g. *sleep like a top*). All potential similes and their components were first extracted using a rule-based algorithm. Then, frozen similes were identified based on reference lists and on the semantic distance between the tenor and the vehicle. The results obtained tend to confirm the fact that frozen similes are not used haphazardly in literary texts. In addition, contrary to how they are often presented, frozen similes often go beyond the ground or the eventuality and the vehicle to also include the tenor.

## 1. INTRODUCTION

Even though literary style is mainly associated with creative writing and deviations from stereotypes, some literary critics have argued that clichés can be used in literary texts for stylistic effects (Amossy & Herschberg-Perrot, 1997). Riffaterre (1964), for example, states that a cliché can either constitute a feature of the author's style that reinforces the literary status of the text or can be used to highlight the moral as well as social behaviours of a certain group of people. According to Abrams (1999), a cliché can be defined as "an expression that deviates enough from ordinary usage to call attention to itself and has been used so often that it is felt to be hackneyed or cloying" (p. 37). This definition definitely echoes the definition that Abrams (1999) gives of the trope: a figure "in which words or phrases are used in a way that effects a conspicuous change in what we take to be their standard meaning" (p. 96). In this respect, it can be said that clichés specifically refer to word combinations that started out as being creative, but, due to their popularity and the passing of time, became phraseological units. Such word combinations include, among others, dead metaphors and cliché similes, which have been vastly studied in phraseology (Wikberg, 2008).

Similes such as La Rochefoucauld's (1) "*Nos actions sont comme des bouts-rimés que chacun tourne comme il lui plaît*" and its translation (2) "*Our actions are like the termination of verses, which we rhyme as we please*" are figures of speech which rely on a linguistic marker to draw a parallel between some explicit or implicit properties that at least two semantically unrelated entities have in common. Since they generally follow the same structure as comparative statements and add concreteness to an utterance by introducing common knowledge, they constitute an inherent part of everyday language. In literary works, similes actively participate in rendering depictions and portrayals real, resonant or even surprising. Moreover, they are flexible enough to be effectively combined with other rhetorical figures such as metaphor, irony, hyperbole or alliteration (Shabat Bethlehem, 1996). Furthermore, as can be seen from the examples (1) and (2), they often follow identical patterns in different languages.

The present study attempts to take advantage of the semantic and syntactic similarities of English and French as far as their simile constructions are concerned in order to, first, extract and mine similes in a corpus of French and British novels written between the 19$^{th}$ and the beginning of the 20$^{th}$ century, and then, determine whether each simile is cliché or not. In addition, it seeks to find out which relevant elements can be used to describe these similes from a corpus-based point of view. In its first section, this paper reviews the rhetorical structure of the similes and computational approaches proposed to identify similes and to account for their creativity. The second section describes the method used to extract as well as to identify cliché similes in written texts. The third section presents the corpora of novels and discusses the results obtained.

## 2. COMPUTATIONAL APPROACHES TO SIMILE IDENTIFICATION

In rhetoric, a simile such as (3) "This girl is graceful like a lily" is made up of the following components:
- a tenor or the object of comparison;
- a *tertium comparationis* or ground: the adjective or verb which denotes the property shared by the compared entities;
- a marker which is the linguistic trigger that introduces the comparison;
- a vehicle which refers to the term that establishes the reference against which the tenor is evaluated (Fishelov, 1993).

In addition to these components, Hanks (2012) also distinguishes the eventuality, the main verb on which the simile is built.

| This girl | is | graceful | like | a lily. |
|-----------|-----------|----------|--------|---------|
| tenor | eventuality | ground | marker | vehicle |

Figure 1. Constituents of the simile "This girl is graceful like a lily"

As far as French is concerned, the comparative clause with an ellipsis of the verb and the adjective of the type "A est B comme C" is generally considered the prototypical form of the simile and has been extensively discussed in rhetoric in relation to the metaphor (Cohen, 1968). If 'comme' is undoubtedly the most frequently used simile marker in French, Shabat Bethlehem (1996) observes that be it in English dictionaries or in research articles, 'like' and 'as' are generally presented as the main or the only simile markers of the English language. This reductionist view of the simile could explain why computational approaches to simile identification have been centred on a rather small number of markers ('like', 'as', 'as ... as', 'more … than', 'less … than').

Automatic simile detection per se can be divided into partial and full simile detection. Partial simile detection has mainly concerned with retrieving specific simile patterns (Veale & Hao, 2007; Veale, Hao & Li, 2008; Wikberg, 2008). It consists either in looking for all corresponding patterns in a corpus or in restricting the search to preselected grounds and vehicles. Moreover, it relies on heuristics or human judgment to differentiate similes from literal comparisons.

In contrast, full simile detection extracts and analyses all sentences containing a simile marker in unrestricted texts, identifies the different components of each potential simile and separates literal similes from figurative ones. As there exists, in comparative statements, a correlation between the syntactic function of a word and its semantic role, Niculae (2013) uses dependency parsing to extract and to identify the components of English similes with 'like'. He also proposes a method to recognise creative similes by measuring the semantic distance between the tenor and the vehicle using distributional semantics. The approach suggested, however, only concerns similes that have both a nominal tenor and a nominal vehicle and requires both nouns to be present in the corpus used to retrieve distributional statistics. In addition, it does not take into account certain syntactic structures such as coordinated verbs or adjectives.

## 3. PROPOSED METHOD

The present study is restricted to nominal similes of the form adjectival ground + simile marker + nominal vehicle (e.g. *cunning as a fox)* and eventuality + simile marker + nominal vehicle (e.g. *cry like a baby*). Since similitude or dissimilitude is primarily inferred by meaning, unlike previous works on cliché similes (Bolshakov, 2003), not only traditional markers, but also other markers which can be combined with these two predefined simile patterns were included. Table 1 lists all selected markers as well as their corresponding simile structures.

The proposed approach to the detection of frozen similes in literary texts can be divided into two main parts:
- extracting and mining of similes, comparisons and pseudo-comparisons;
- and filtering all the extracted elements based on their idiomaticity and the semantic distance between the vehicle and the tenor.

|  | **English** | **French** | **Possible structures** |
|---|---|---|---|
| *Prepositions and adverbs* | like, unlike, as, as…as, more…than, less…than, -er … than | comme, ainsi que, de même que, autant que, plus…que, tel que, moins…que, aussi…que | - Verb + marker + vehicle<br>- Tenor + verb + marker + vehicle<br>- Adjectival ground + marker + vehicle<br>-Tenor + adjectival ground + marker + vehicle<br>-Tenor + verb + adjectival ground + marker + vehicle |
| *Prepositional phrases* |  | à l'image de, à l'instar de, à la manière de, à l'égal de, à la manière de, à la façon de |  |

Table 1. Similes markers for English and French

## 3.1. The simile extraction and mining module

The different steps detailed in this section are derived from the rule-based algorithm for simile mining presented in Mpouli and Ganascia (2015). The extraction and mining phase comprises various preprocessing tasks, simile candidate sentence extraction and finally, the identification of the components of each extracted simile candidate. Preprocessing tasks include tokenisation, part-of-speech tagging, syntactic chunking and sentence segmentation. The first three tasks are performed with TreeTagger (Schmid 1994), a freely available multilingual tokeniser, part-of-speech tagger and chunker[1]. The sentence boundaries defined by TreeTagger constitute the first basis for sentence segmentation and are corrected with specific rules when an ellipsis, a question or an exclamation mark is not followed by a capital letter.

```
Guests, like fish, begin to smell after three days.

<NC> Guests (NNS) </NC> , <PC> like (IN) <NC> fish (NN) </NC>
</PC> , (,) <VC> begin (VVP) </VC> <VC> to (TO) smell (VV)
</VC> <PC> after (IN) <NC> three (CD) days (NNS) </NC> . (SENT)

    NC = noun chunk, PC = prepositional chunk, VC = verb chunk
```

Figure 2. Example of a chunked sentence

Syntactic chunking is an intermediary stage between part-of-speech tagging and constituency or dependency parsing and produce "flat, non-overlapping segments of a sentence that constitute the basic non-recursive phrases corresponding to the major parts-of-speech found in most wide-coverage grammars" (Jurafsky & Martin, 2009, p. 485). These chunks combined with handcrafted rules are essential for the next two phases. Since chunks do not give information about the grammatical function of a word, the algorithm mainly relies on syntax, dependency grammar and syntactic clues. For example, based on the syntactic order prevalent in English and in French, it can be deduced that the vehicle would be the head noun of the noun phrase that follows the

---
[1] http://www.cis.uni-muenchen.de/~schmid/tools/TreeTagger/

marker either directly or after an appositive phrase. In addition, the algorithm takes into consideration the ambiguity inherent to some comparative constructions and, depending on the sentence structure, labels all words that can plausibly be a component of the simile. Consequently, in a sentence such as "[...] a spark was kindled that wanted but opportunity to blaze into a flame, pure and bright as the shrine on which it burned", the sentence is analysed as follows:
- marker: 'as'
- vehicle: 'shrine'
- grounds: 'pure', 'bright' → tenor: 'flame'
- eventuality: 'blaze' → tenor: 'spark'

Table 2 summarises the different characteristics and textual clues used to identify each potential simile component.

| Constituent | Grammatical category | Informative Clues | Governor |
|---|---|---|---|
| Adjectival ground | Adjective, past or present participle | Not separated from the marker by a coordinating conjunction, a relative pronoun, a preposition or a noun phrase | / |
| Tenor – Noun that the adjectival ground modifies | Common noun | Part of the noun phrase before or after the adjective | Non-predicative adjectival ground |
| Tenor – postposed direct object | Common noun | Not after a preposition Follows a verb or a prepositional phrase that follows a verb | Verb |
| Tenor – preposed direct object | Common noun | Part of the noun phrase directly before 'que', 'that', 'which' and the subject | Verb |
| Tenor – objective personal pronoun (direct object) | Personal pronoun | Directly before a verb | Verb |
| Tenor – subjective personal or demonstrative pronoun | Personal and demonstrative pronouns | Directly before or after a conjugated verb | Verb |
| Tenor – subject | Common or proper noun | Before a verb and not after a preposition | Verb |
| Eventuality | Verb | Not separated from the marker by a colon or a semi-colon | / |
| Vehicle – common noun | Common noun | Head of a noun phrase Not separated from a verb that follows him by a punctuation mark, a relative pronoun subject, a subjective personal pronoun, a coordinating or subordinating conjunction | Marker |

Table 2. Correlation between each type of constituent, the clues to identify it and its grammatical function

## 3.2. Detection of frozen similes

In order to detect frozen similes, all ground/eventuality + vehicle couples were first compared to a list of frozen similes of the corresponding language, compiled from different sources such as *Les Comparaisons du français* by Nicolas Cazelles (1996) or *Dictionnaire français/anglais des comparaisons* by Michel Parmentier (2002). Since the resulting reference lists of frozen similes were mostly based on sources that were not contemporary with the novels of the corpus and were by no means comprehensive, frequency and semantic distance were used to single out frozen similes among the remaining ground/eventuality + vehicle couples. All ground/eventuality + vehicle couples that appeared at least 5 times in novels by different authors were thus selected. Then, the semantic distance between the tenor and the vehicle in each sentence was assessed.

When both the tenor and the vehicle were nouns, their semantic categories were extracted from two machine-readable dictionaries: Wordnet (Fellbaum, 1998) and *Le Dictionnaire électronique des mots* (Dubois & Dubois-Charlier, 2010) for English and French respectively. As far as Wordnet is concerned, the number of the noun's corresponding lexicographer file, which precedes the term in every definition, was taken as its semantic category. For *Le Dictionnaire électronique des mots,* three semantic categories are provided: 'animal' and 'humain' (human being) and 'non-animé' (inanimate).

Semantic distance, however, required manual disambiguation in the following cases:
- when different semantic categories were attributed to either the tenor or the vehicle;
- when an English tenor or vehicle was classified as a unique beginner for nouns;
- when both French nouns were tagged as 'non-animé';
- when the tenor was a personal pronoun.

| My cousin behaves like a child.<br>09972010 **18** n 04 cousin<br>09918248 **18** n 02 child<br>18   Noun denoting people | Sa voix était aussi faible qu'un souffle.<br>- voix: type="non-anime"<br>- souffle: type="non-anime" |
|---|---|

Figure 3.  Semantic categories extracted from Wordnet (left) and *Le Dictionnaire électronique des mots* (right)

## 4. RESULTS AND DISCUSSION

## 4. 1. Presentation of the corpus

The two corpora used for this experiment were built with digital versions of literary texts in the public domain, collected mainly from the Project Gutenberg website[2] and from the Bibliothèque électronique du Québec[3], for British and French novels respectively. Most of the novels included in the corpus were written during the 19th century so as to ensure linguistic homogeneity and because that century witnessed the novel imposing itself as a predominant literary genre. In addition, a ratio of at least 3 novels per writer was observed and the novels were not restricted to a specific literary genre. This method enabled to create a corpus of 1,191 British novels authored by 62 writers and a corpus of 746 French penned by 57 novelists. In terms of size, the British corpus contains 152,941,750 tokens and its French counterpart, 119,914,914 tokens.

---
[2] www.gutenberg.org
[3] beq.ebooksgratuits.com

### 4.2. Description of detected frozen similes

The rather low number of total occurrences of the top frozen similes in each language (listed in Table 3) seems to confirm the fact that clichés in general and frozen similes in particular are far from being common in literary texts. In addition, French novelists are more inclined to use frozen similes than British ones. One striking result of this study is the fact that the same simile (*pale + marker + death* and *pâle + marker + mort*) is the most frequently used in both languages. This simile is also interesting in itself, from a stylistic point of view, as it assigns a human feature to an abstract entity. As a matter of fact, 'pale' and its synonym, 'white', appear several times in the most frequent similes. Similarly, 'death' is used in three of the top frozen English similes. Apart from the fact that 'paleness' mostly characterises humans, its context of occurrence suggests that it is generally used to stress somebody's distress or fear in reaction to a particular news or event. It also sometimes conveys the impression that the narrator has of a protagonist. Generally speaking, this simile is used to produce maximal effect, such as in: "The old Cavalier looked pale as death, and greatly agitated".

| English | French |
|---|---|
| *pale + marker + death (152)* | *pâle + marker + mort (283)* |
| *cold + marker + ice (128)* | *pleurer + marker + enfant (188)* |
| *bad + marker + death (114)* | *immobile + marker + statue (179)* |
| *white + marker + death (108)* | *rapide + marker + éclair (164)* |
| *good + marker + gold (108)* | *blanc + marker + neige (162)* |
| *white + marker + sheet (102)* | *aimer + marker + frère (140)* |
| *good + marker + word (98)* | *tomber + marker + massue (135)* |
| *come + marker + shock (80)* | *tuer + marker + chien (122)* |
| *black + marker + night (87)* | *pâle + marker + morte (121)* |
| *white + marker + snow (83)* | *beau + marker + ange (115)* |
| *silent + marker + grave (83)* | *passer + marker + éclair (112)* |
| *clear + marker + daylight (83)* | *rapide + marker + pensée (106)* |

Table 3. The 12 most frequent frozen similes in both corpora

Furthermore, several detected frozen similes, in fact, are variants of another frozen simile. Frequency could therefore help to determine the main simile that the others seek to replace. For example, 'motionless + marker + statue', 'rigid + marker + statue', 'motionless + marker + image' are all non-idiomatic variations of 'immobile + marker + statue'.

Still based on the frequency of frozen similes in each corpus, it is possible to propose a scale of clichédness that can further describe similes for stylistic analysis. For instance, 'white as the snow" would be a prominent literary cliché, while 'heavy + marker + lead' (42 occurrences) would be a medium literary cliché and 'harmless + marker + dove' (6 occurrences) would be a relatively rare literary cliché.

As far as compositionality is concerned, some frozen similes use various markers while others always use the same one, especially when they make use of the comparative forms "more/less… than" or when they are derived from proverbial expressions such as "blood is thicker than water". In addition, some frozen similes tend to always be associated with the same tenor or different tenors belonging to the same semantic fields. Examples of this type of similes include 'eyes + wide + saucers', 'cheveu +noir + aile de corbeau', ' nez +recourbé + bec' and 'yeux + brûler/briller + charbon'. In this respect, it is possible to say that some literary frozen similes are restricted to specific cognitive associations, so much so that they come automatically in mind when one wants to emphasise a particular state or attribute of an entity.

## 5. CONCLUSION

The aim of this paper was to study frozen literary similes in a corpus of novels written in English and in French. If the greater part of the extraction of similes and of their components was done automatically, the recognition of frozen similes still relies partially on human knowledge. In this respect, it seems necessary for future work to research how to disambiguate first the tenor or the vehicle so as to find directly its corresponding semantic category. Since this study is mostly oriented towards a stylistic description of similes, another perspective would be to add information about the literary clichédness of a particular simile to a simile mining system, in order to shade new light on the perception and interpretation of frozen similes in literature as a whole.

**Acknowledgment**

This work was supported by French state funds managed the ANR within the Investissements d'Avenir programme under the reference ANR-11-IDEX-0004-02.